\documentclass[manuscript,sigconf]{acmart}

\usepackage{multirow}
\usepackage{makecell}

\AtBeginDocument{%
  }

\setcopyright{none}

\renewcommand\footnotetextcopyrightpermission[1]{}

\begin{document}
\begin{sloppypar}

\title{Stage-Diff: Stage-wise Long-Term Time Series Generation Based on Diffusion Models}

\author{Xuan Hou}
 \affiliation{
   \institution{Shandong University}
   \city{Qingdao}
   \country{China}
 }
    \email{houxuan@mail.sdu.edu.cn}
 
\author{Shuhan Liu}
 \affiliation{
   \institution{Shandong University}
      \city{Qingdao}
   \country{China}
 }
    \email{shuhanliu@mail.sdu.edu.cn}
 
 \author{Zhaohui Peng}
 \affiliation{
   \institution{Shandong University}
      \city{Qingdao}
   \country{China}
 }
  \email{pzh@sdu.edu.cn}
 \authornote{Corresponding author.}
 
 \author{Yaohui Chu}
 \affiliation{
   \institution{Shandong University}
      \city{Qingdao}
   \country{China}
 }
     \email{cyh0206@mail.sdu.edu.cn}
     
 \author{Yue Zhang}
 \affiliation{
   \institution{Shandong University} 
      \city{Qingdao}
   \country{China}
   }
   \email{zhangyue_zz@mail.sdu.edu.cn}
   
 \author{Yining Wang}
 \affiliation{
   \institution{Shandong University}
      \city{Qingdao}
   \country{China}
 }
     \email{wangyning@mail.sdu.edu.cn}
     


\renewcommand{\shortauthors}{Hou X., Liu S. et al.}

\begin{abstract}
  Generative models have been successfully used in the field of time series generation. However, when dealing with long-term time series, which span over extended periods and exhibit more complex long-term temporal patterns, the task of generation becomes significantly more challenging. Long-term time series exhibit long-range temporal dependencies, but their data distribution also undergoes gradual changes over time. Finding a balance between these long-term dependencies and the drift in data distribution is a key challenge. On the other hand, long-term time series contain more complex interrelationships between different feature sequences, making the task of effectively capturing both intra-sequence and inter-sequence dependencies another important challenge. To address these issues, we propose \textbf{Stage-Diff}, a staged generative model for long-term time series based on diffusion models. First, through stage-wise sequence generation and inter-stage information transfer, the model preserves long-term sequence dependencies while enabling the modeling of data distribution shifts. Second, within each stage, progressive sequence decomposition is applied to perform channel-independent modeling at different time scales, while inter-stage information transfer utilizes multi-channel fusion modeling. This approach combines the robustness of channel-independent modeling with the information fusion advantages of multi-channel modeling, effectively balancing the intra-sequence and inter-sequence dependencies of long-term time series. Extensive experiments on multiple real-world datasets validate the effectiveness of Stage-Diff in long-term time series generation tasks.
\end{abstract}




\maketitle

\section{Introduction}

\begin{figure}[htbp]
  \centering
  \includegraphics[width=\linewidth]{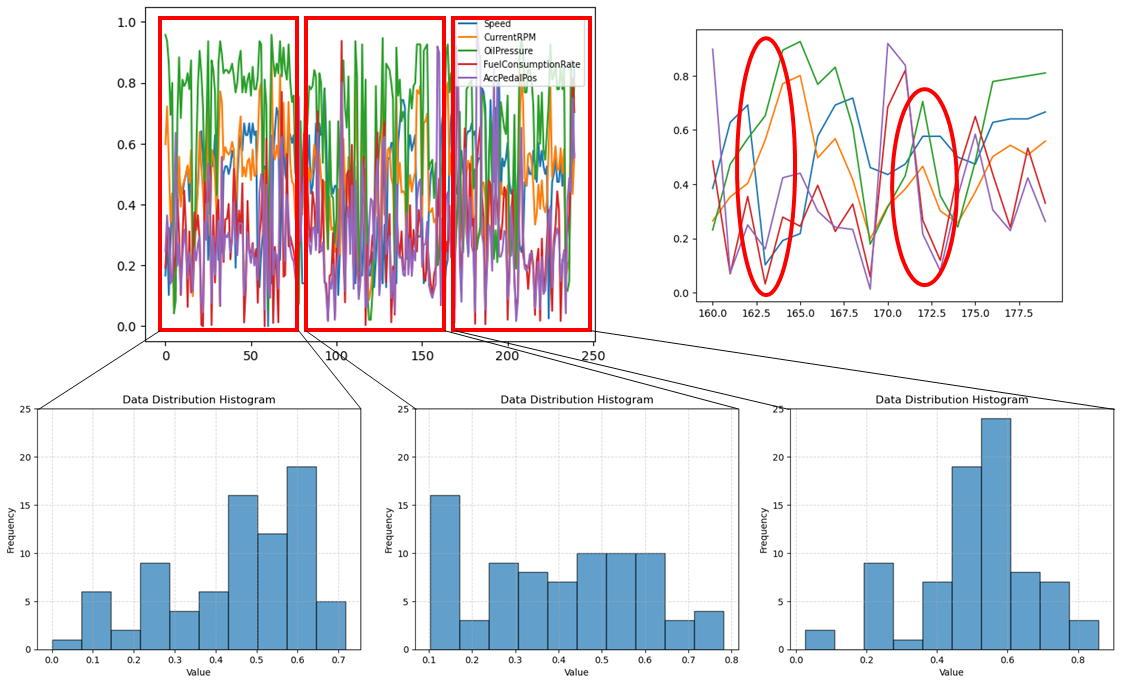}
  \caption{State change sequence diagram of commercial vehicle}
  \label{fig:fig1}
\end{figure}

In recent years, with the rapid advancement of generative model techniques, time series generation has become a significant research focus, drawing widespread attention in areas such as data augmentation, forecasting, and missing value imputation\cite{1-timegan} \cite{2} \cite{3} \cite{4}. However, when dealing with long-term time series that span extended durations and exhibit more complex temporal patterns, the generation task faces greater challenges.

Long-term time series often exhibit long-term dependencies, meaning that the state at the current time step may be closely linked to the states at multiple past time steps. Furthermore, over extended time periods, the data distribution of the time series tends to undergo gradual changes. As shown in Figure \ref{fig:fig1}, which illustrates the changes in several state variables of a commercial vehicle over time, the three data distribution histograms below reveal noticeable differences in the data distributions across different time periods. Overemphasizing the long-term trend of the time series may lead to the neglect of these distribution shifts. Therefore, finding a proper balance between long-term temporal dependencies and the gradual changes in data distribution is a key challenge in long-term time series generation.

Furthermore, long-term time series exhibit more complex inter-sequence dependencies. Many deep learning-based methods for long-term time series forecasting typically focus on capturing temporal dependencies, but combine varied series into a unified hidden-time embedding, without explicitly modeling the relationships between the sequences. Recent research in time series forecasting has found that decomposing multivariate time series into multiple univariate time series can actually improve the long-term forecasting performance\cite{5} \cite{6}. While this approach avoids interference from the complex relationships between different sequences and enhances the robustness of the forecasting model, it clearly results in a loss of the ability to capture these inter-sequence dependencies. As shown in Figure \ref{fig:fig1}, the changes between different sequences often exhibit both independence and interdependence. Therefore, effectively capturing both intra-sequence and inter-sequence dependencies remains a significant challenge in long-term time series generation.

To address the aforementioned challenges, we propose Stage-Diff, a staged long-term time series generation model based on diffusion models. This model divides the generation of long-term time series into staged generation and inter-stage information transfer. First, inter-stage information transfer ensures consistency in long-term temporal dependencies, while the staged generation approach facilitates the modeling of data distribution shifts in long-term time series. Second, during the modeling process of each stage, progressive sequence decomposition based on attention mechanisms is employed to extract univariate time series trend information incrementally across different time scales. Subsequently, in the inter-stage information transfer process, multi-channel information fusion modeling is applied to capture correlations among trend information from different time series at varying time scales. This fused information is then used as hidden historical information for the subsequent stage, effectively capturing both intra-sequence and inter-sequence dependencies. The main contributions of this chapter are as follows:

\begin{itemize}
    \item We propose Stage-Diff, a staged long-term time series generation method based on diffusion models. By employing staged generation and inter-stage information transfer, the method ensures the long-term dependency of time series while effectively addressing the modeling of data distribution shifts.
    \item Through progressive sequence decomposition, Stage-Diff performs channel-independent modeling across different time scales and employs multi-channel information fusion during inter-stage information transfer. This approach effectively balances intra-sequence and inter-sequence dependencies in long-term time series.
    \item Extensive experiments on multiple real-world datasets have been conducted to validate the effectiveness of the proposed Stage-Diff model, and ablation studies further demonstrate the contribution of key modules.
\end{itemize}

\section{Related Work}
\subsection{Time Series Generation}
Early generation models relied mainly on traditional statistical methods, which generated synthetic time series datasets by constructing approximate distributions and sampling from them\cite{7} \cite{8}. However, because these approximate distributions often failed to accurately capture the characteristics of real data, the quality and diversity of the generated time series data were significantly limited. With the rapid advancement of deep learning technologies, time series generation methods based on deep generative models, such as Generative Adversarial Networks (GANs)\cite{1-timegan} \cite{9} \cite{10} \cite{11} \cite{12} and Variational Autoencoders (VAEs)\cite{13-timevae} \cite{14}, have garnered widespread attention.

In recent years, diffusion models\cite{15} have achieved groundbreaking progress in image generation\cite{16} \cite{17}, demonstrating their powerful potential in generative tasks. This success has inspired researchers to adapt the concepts and techniques of diffusion models to the field of time series generation. Consequently, the application of diffusion models in time series generation has emerged as a popular and cutting-edge research direction\cite{18} \cite{19} \cite{20}.

\subsection{Time Series Modeling}
Early time series modeling primarily relied on Recurrent Neural Networks (RNNs) and Convolutional Neural Networks (CNNs). However, RNNs tend to suffer from long-term memory loss when handling long-term time series, while CNNs are constrained by their limited receptive fields, making it challenging to capture long-term dependencies in time series. These limitations hinder the effectiveness of traditional methods in modeling the complex temporal dependencies of long-term time series.

In recent years, Transformer\cite{21} has become a popular research focus in the field of time series modeling due to its superior performance in capturing global correlations. A plethora of Transformer-based methods have emerged for long-term time series modeling\cite{22} \cite{23} \cite{24} \cite{25} \cite{26}. However, a recent study\cite{5} questioned the applicability of Transformer in time series modeling, revealing that a simple linear model outperformed all state-of-the-art Transformer-based approaches. This finding sparked widespread discussion and motivated subsequent research to improve Transformer models for time-series tasks. These advancements demonstrate that Transformer still holds significant potential in effectively capturing the complex patterns of time series, particularly in handling long-term dependencies\cite{6} \cite{27} \cite{28}.

\begin{figure*}[htbp]
  \centering
  \includegraphics[width=0.9\textwidth]{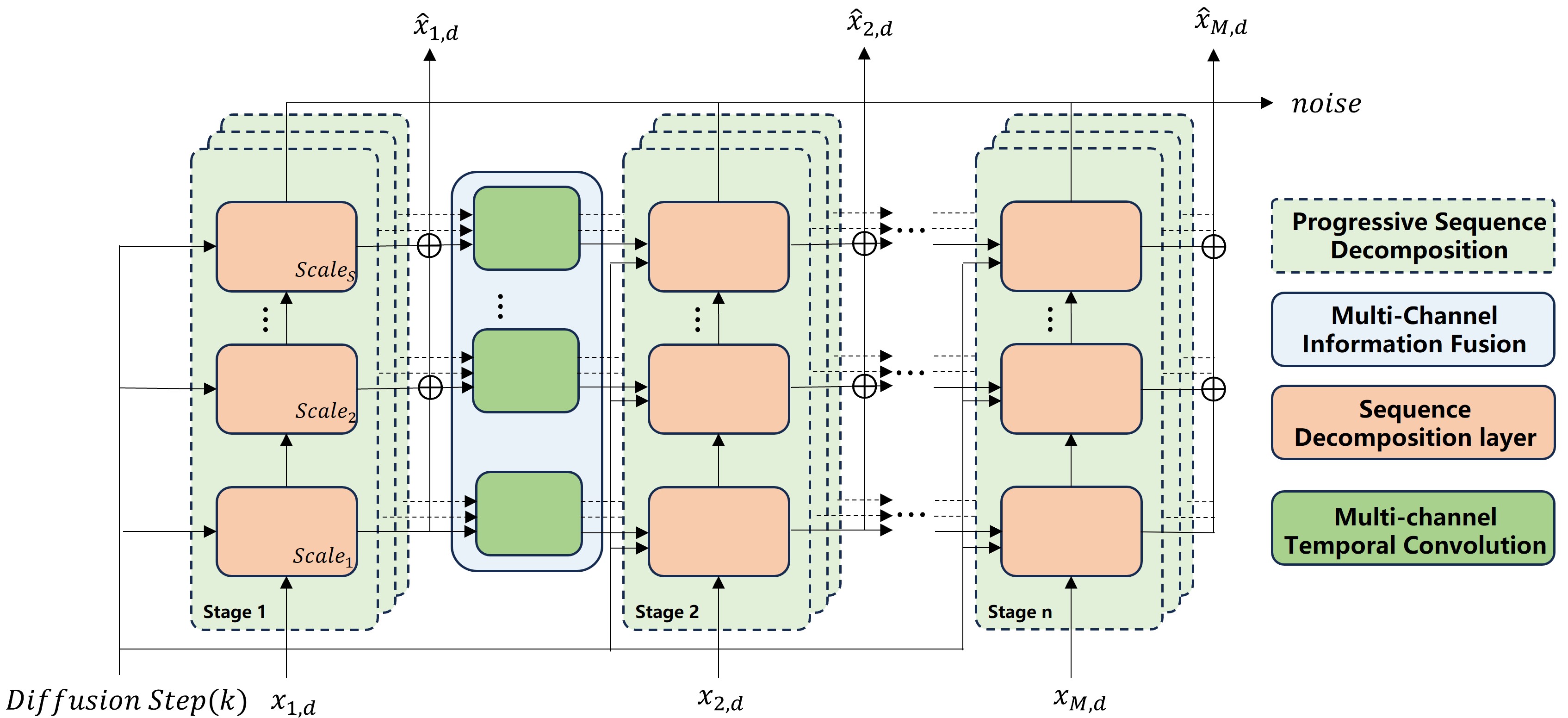}
  \caption{The framework of the Stage-Diff}
  \label{fig:model}
\end{figure*}

\section{Methodology}
Stage-Diff primarily adopts a staged time series generation approach, which includes a progressive sequence decomposition module for handling individual time series stages and a multi-channel information fusion module for inter-stage information transfer, as shown in Figure \ref{fig:model}. The staged decomposition effectively addresses the issue of data distribution shifts in long-term time series, while inter-stage information transfer enables the preservation of long-term temporal dependencies. Within the progressive sequence decomposition module, a channel-independent modeling approach is adopted. A Transformer-based sequence encoder-decoder is utilized to incorporate historical information while performing sequence decomposition, progressively extracting trend information and residual information at different temporal scales for the current time stage. The multi-channel information fusion module integrates multi-scale temporal trend information extracted from different stages and uses it as historical information for input into the next stage. This approach strikes a balance between capturing both intra-sequence and inter-sequence dependencies.

We use $X_{1:{L_{ser}}}=({X_1,X_2,\cdots,X_{L_{ser}}}) \in R^{{L_{ser}} \times D} $to represent a time series of length $L_{ser}$, $D$ is the feature dimension of the time series. Given the dataset $DA= \{X^i_{1:L_{ser}}\}_{i=1}^{N}$ contains $N$ samples of time-series, our goal is to use a diffusion-based generator to approach the function $\widehat{X}^i_{1:L_{ser}}=G(Z_i)$ which maps random sequence $Z_i \in R^{L_{ser} \times D}$ to the target sequence space. As mentioned earlier, Stage-Diff adopts a staged generation approach. For an input long-term time series $X_{1:{L_{ser}}}$, it is divided into $M$ satges time-series $\{x_m \}_{m=1}^M$, each with a length of $L_{sta}$ serving as the input for different time stages, where $M \times L_{sta} = L_{ser}$.

\subsection{Diffusion Framework}

\begin{figure*}[htbp]
  \centering
  \includegraphics[width=0.85\textwidth]{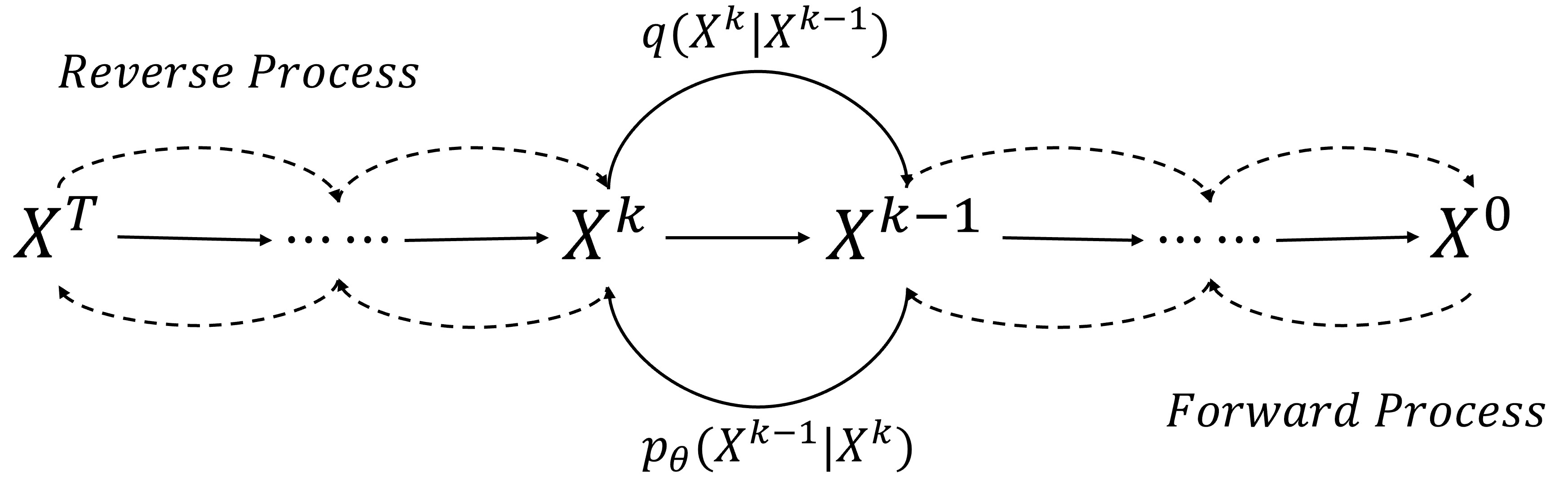}
  \caption{The Framework of Diffusion Model}
  \label{fig:diff}
\end{figure*}

We first introduce the overall framework of the diffusion model, as shown in the Figure\ref{fig:diff}. This model is a type of generative model based on probabilistic diffusion processes, which generates high-quality synthetic samples by progressively learning the reverse diffusion process of the data distribution. Its core framework consists of two main processes: the forward process and the reverse process. In the forward process, samples $X^0 \sim q(X)$ from the real data distribution are gradually noised, generating a series of samples $X_1, X_2, \cdots, X_T$ that approach a Gaussian distribution. The process is defined as:
\begin{equation}
  q(X^k|X^{k-1})=\mathcal N (X^k;\sqrt{\alpha_k} X^{k-1}, (1-\alpha_k)I),
\end{equation}
Here, $\alpha_k \in (0,1)$ is a hyperparameter that controls the amount of noise added at each step, and it typically decreases over time. After several iterations, $X^T$ will ultimately approach a Gaussian distribution. Using the reparameterization technique, the sample distribution at any time step can be directly computed from $X^0$. The formula is as follows:
\begin{equation}
  q(X^k|X^0)= \mathcal N (X^k;\sqrt{1-\beta_k} X^{0}, \beta_k I),
\end{equation}
here, $\beta_k=1-\Bar{\alpha}_k$, $\Bar{\alpha}_k=\sum^k_{i=1}\alpha_i$. At this point, $X^k$ can be expressed as:
\begin{equation}
  X^k = \sqrt{1-\beta_k}X^0+\sqrt{\beta_k}\epsilon,
\end{equation}
where $\epsilon$ is the noise sampled from the standard Gaussian distribution $\mathcal N(0, I)$. This equation also allows for the reconstruction of $X^0$ from $X^k$.
The reverse diffusion process is a Markov process that iteratively removes noise, generating realistic data from random noise. In the $k_th$ denoising step, $X^{k-1}$ is sampled from the following normal distribution using $X^{k}$:
\begin{equation}
  p_\theta(X^{k-1}|X^K) = \mathcal N({X^{k-1}; \mu_\theta(X^k,k), \sigma_\theta(X^k,k)}),
\end{equation}
here, the variance $\sigma_\theta(X^k,k)$ is typically fixed as $\frac{1-\alpha_k}{\alpha_k}$, and the mean $\mu_\theta(x^k, k)$ is usually defined by a neural network with parameters $\theta$. Therefore, this problem is generally formulated as a noise estimation or data prediction problem. For the noise estimation problem, the neural network $\epsilon_\theta$ predicts the noise added to the diffusion input $X^k$. In this case, the mean $\mu_\theta(x^k,k)$ can be obtained from the following equation:
\begin{equation}
  \mu_\theta(X^k,k)=\frac{1}{\sqrt{\alpha_k}}X^k-\frac{1-\alpha_k}{\sqrt{1-\Bar{\alpha}_k}\sqrt{\alpha_k}}\epsilon_\theta(X^k,k),
\end{equation}
The parameters $\theta$ can be optimized by $L_\epsilon=E_{k,x^0,\epsilon}[{||\epsilon-\epsilon_\theta(X^k,k)||}^2]$. For the data prediction problem, a denoising neural network $x_\theta$ is used to obtain an estimate of the clean data $X^0$ given the diffusion input $X^k$, denoted as $x_\theta(X^k,k)$. In this case, the mean value $\mu_\theta(X^k,k)$ can be obtained using the following equation:
\begin{equation}
  \mu_\theta(X^k,k)=\frac{\sqrt{\alpha_k}(1-\Bar{\alpha}_{k-1})}{1-\Bar{\alpha}_k}X^k-\frac{\sqrt{\Bar{\alpha}_{k-1}}(1-\alpha_k)}{1-\Bar{\alpha}_k}x_\theta(X^k,k),
\end{equation}
The parameters $\theta$ can be optimized by $L_x=E_{x^0,\epsilon,k}[{||X^0-x_\theta(X^k,k)||}^2]$.

The Stage-Diff model proposed in this paper adopts the aforementioned data prediction approach to restore clean data from noised data. For convenience, in the following model description, we will refer to the denoising process at the $k_th$ diffusion step without explicitly labeling the diffusion steps.

\subsection{Progressive Sequence Decomposition Module}
The progressive decomposition module is to model the stage time series with a single channel. For stage $m$, the stage time series $x_m \in R^{D \times L_{sta}}$ consists of $D$ feature sequences, which are input into $D$ progressive sequence decomposition modules. The input for each module is $X_{m,d} \in R^{1\times L_{sta}}$, where $d=\{1,2,\dots,D\}$. Inside the module, sequential decomposition layers are stacked to progressively decompose the time series and capture trend information and residual information at different time scales. The residual information is passed to the next layer for further decomposition, while the trend information from different scales is summed to estimate the clean data for the stage $\hat{x}_{m,d}$.

\begin{figure}[htbp]
  \centering
  \includegraphics[width=\linewidth]{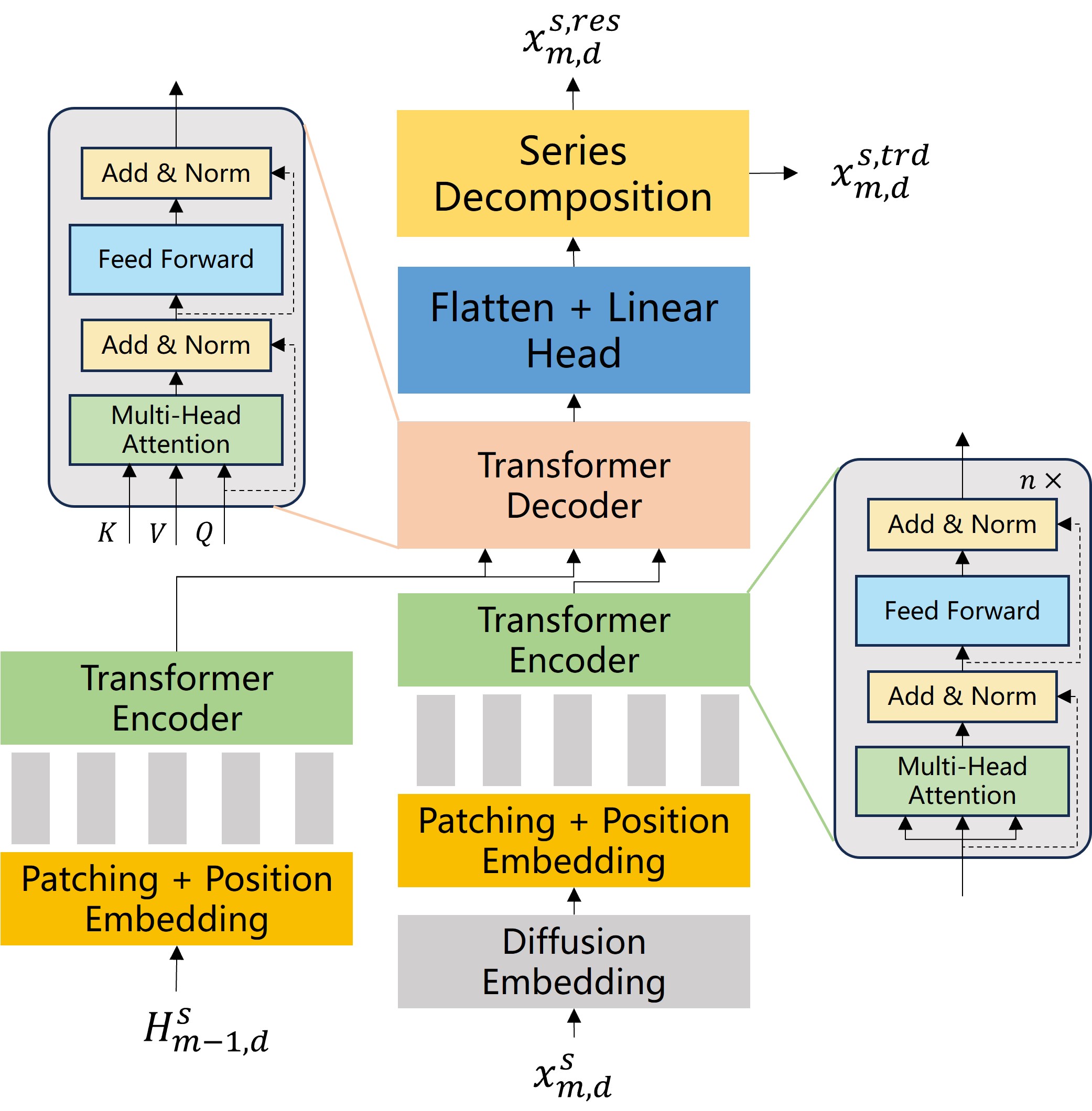}
  \caption{Series Decomposition Layers}
  \label{fig:seriesDecom}
\end{figure}

Taking the $s\mbox{-}th$ time scale as an example, the sequence decomposition layer, as shown in the Figure \ref{fig:seriesDecom}, mainly includes diffusion embedding, patching and positional encoding, Transformer encoder, Transformer decoder and sequence decomposition operation.

\textbf{Diffusion Embedding}. For the input time series, it is necessary to perceive the current diffusion step, so diffusion embedding is required. We use the classical sinusoidal position embedding. Since this embedding process does not alter the data dimensions, no additional symbolic annotation is needed.

\textbf{Patching and Position Embedding}. Time series exhibit strong local dependencies. The dynamic changes over time reflect the relationships along the time dimension, while the values at individual time steps often lack rich temporal semantics. Therefore, for the input univariate time series $x^s_{m,d}$, a Patching operation is applied, where a sliding window of length $L_{patch}$ divides the series into $P$ time segments, with $P=\frac{L_{sta}-L_{patch}}{L_{win}}+1$, where $L_{win}$ is the sliding step size. The attention mechanism shortens the reasoning path for inputs at different positions to $1$, effectively capturing the long-term dependencies in the time series. However, this also leads to the loss of positional information, which is crucial for time series, as the relationship between the input's past and future is highly significant. Therefore, before the input is passed into the Transformer encoder, positional encoding is applied. A trainable linear projection $W_{p} \in R^{D\times L_{patch}}$ is used to project the Patch sequence into the Transformer latent space of dimension $D$, and a learnable positional encoding $W_{pos} \in R^{D \times P}$ is used to preserve the positional relationships of the input sequence. This can be expressed as:
\begin{equation}
  x_{m,d}^{s,p} = W_p (Patching(x_{m,d}^{s}))+W_{pos}.
\end{equation}
\textbf{Transformer Encoder}. The position-enhanced Patch sequence is further processed through the Transformer encoder to extract contextual information. The core operation lies in the multi-head attention mechanism. For each attention head $h=1,2,\cdots,H$, the input is mapped into the query matrix $Q_h^d = W_h^Q x_{m,d}^{s,p}$, key matrix $K_h^d = W_h^K x_{m,d}^{s,p}$, and value matrix $V_h^d=W_h^V x_{m,d}^{s,p}$, where $W_h^Q, W_h^K\in R^{d_k \times D}$, $W_h^V \in R^{D \times D}$ are learnable weights. Then, scaled production is used to obtain the attention output $O_h^d \in R^{D \times P}$:
\begin{equation}
  (O_h^d)^T = Attention(Q_h^d, K_h^d, V_h^d),
\end{equation}
\begin{equation}
  Attention(Q, K, V)=Softmax(\frac{(Q)^T K}{\sqrt{d_k}})(V)^T,
\end{equation}
The multi-head attention mechanism also includes a batch normalization layer and a feedforward network with residual connections, as shown in the Figure\ref{fig:seriesDecom}. The resulting output is denoted as $z^d \in R^{D \times P}$.
The historical input $H^s_{m-1,d}$ is processed through the same network structure, as shown in the Figure\ref{fig:seriesDecom}, and the resulting output is denoted as $z_{his}^d \in R^{D \times P}$.

\textbf{Transformer Decoder}. The decoder integrates multi-channel historical information, ensuring the long-term temporal dependencies of the generated time series. The structure of the Transformer decoder is consistent with that of the Transformer encoder, with the key difference lying in its input. Instead of processing a single sequence, the decoder takes both the historical sequence $z_{his}^d$ and the encoder-enhanced current sequence  $z^d$ as inputs. Using learnable weight matrices, $z_{his}^d$ is mapped to the key matrix $K$ and value matrix $V$, while $z^d$ is mapped to the query matrix $Q$. Historical information is fused through the attention mechanism. Finally, a linear-headed unfolding layer produces the sequence representation at this time scale $\hat{x}_{m,d}^{s} \in R^{1\times L_{sta}}$.
\textbf{Series Decomposition}. TTo learn complex long-term temporal patterns, this module adopts a layer-by-layer decomposition approach, sequentially decomposing the time series across multiple time scales. At each scale, the time series is decomposed into trend information and residual information. The trend information represents the long-term variations of the time series at the given time scale, while the residual information captures the short-term fluctuations and noise after removing the trend. The residual information is further decomposed in the subsequent decomposition layers, enabling the model to begin with simple long-term trends and progressively capture finer-grained short-term variations. This approach allows the model to focus on features across different time scales. The specific decomposition operation is expressed as follows:
\begin{equation}
  x_{m,d}^{s,trd} = AvgPool(Padding(\hat{x}_{m,d}^{s})),
\end{equation}
\begin{equation}
  x_{m,d}^{s,res} = \hat{x}_{m,d}^{s} - x_{m,d}^{s,trd},
\end{equation}
Here, $x_{m,d}^{s,trd},x_{m,d}^{s,res} \in R^{1 \times L_{sta}}$ represent the trend information and residual information extracted at this time scale, respectively. An average pooling operation $AvgPool(\cdot)$ with sequence padding $Padding(\cdot)$ is used to maintain the sequence length.  The residual information $x_{m,d}^{s,res}$ serves as the input $x^{s+1}_{m,d}$ for the next time scale's sequential decomposition layer. The fusion of trend information across all time scales provides the estimated clean time series for the current channel at this time stage, expressed as $\hat{x}_{m,d} = \frac{1}{S}\sum_{s=1}^{S}x_{m,d}^{s,trd}$.


\subsection{Multi-Channel Information Fusion Module}
The multi-channel information fusion module takes the trend information from different time scales, output by the progressive sequence decomposition module in the previous stage, and performs information fusion before passing it to the next stage to ensure the long-term dependency of the time series. Since single-channel modeling is used within each stage, interactions between different channels are ignored, which is often a critical factor leading to long-term time series distribution shift. To address this, the module employs multi-channel modeling. To maintain consistency across time scales, this module uses multi-channel temporal convolution networks for information fusion at each time scale.Taking the 
$s\mbox{-}th$ time scale as an example, the input to the multi-channel temporal convolution consists of the historical trend information $x_{m,d}^{s,trd}$ from $D$ channels, where $d=\{1,2,\cdots,D\}$ represents the feature dimensions. First, the historical trend information from different channels is concatenated into $x_{m,:}^{s,trd} \in R^{D \times L_{sta}} $. Then, the multi-channel temporal convolution is applied to obtain the fused historical information $H_{m,:}^{s} \in R^{D \times L_{sta}}$, represented as:
\begin{equation}
  H_{m,:}^{s}=conv(Padding(x_{m,:}^{s,trd})),
\end{equation}
Here, $Padding(\cdot)$ is a padding operation that ensures the consistency of data dimensions before and after the convolution, and $conv(\cdot)$ represents the multi-channel temporal convolution. The convolution kernel size is $D \times L_{conv}$, $D$ is the feature dimension of the input time series, ensuring the ability to capture the information dependencies across all channels, and $L_{conv}$ is a hyperparameter representing the convolutional window size. The number of output channels of the convolution is also D. $H_{m,:}^{s}$ is then split into multiple single-channel sequences $H_{m,d}^{s}$, where $d=\{1,2,\cdots,D\}$. These single-channel sequences are input to the progressive sequence decomposition module of the next time stage to assist in the decomposition of time series.

\begin{figure*}[hbtp]
  \centering
  \includegraphics[width=0.9\textwidth]{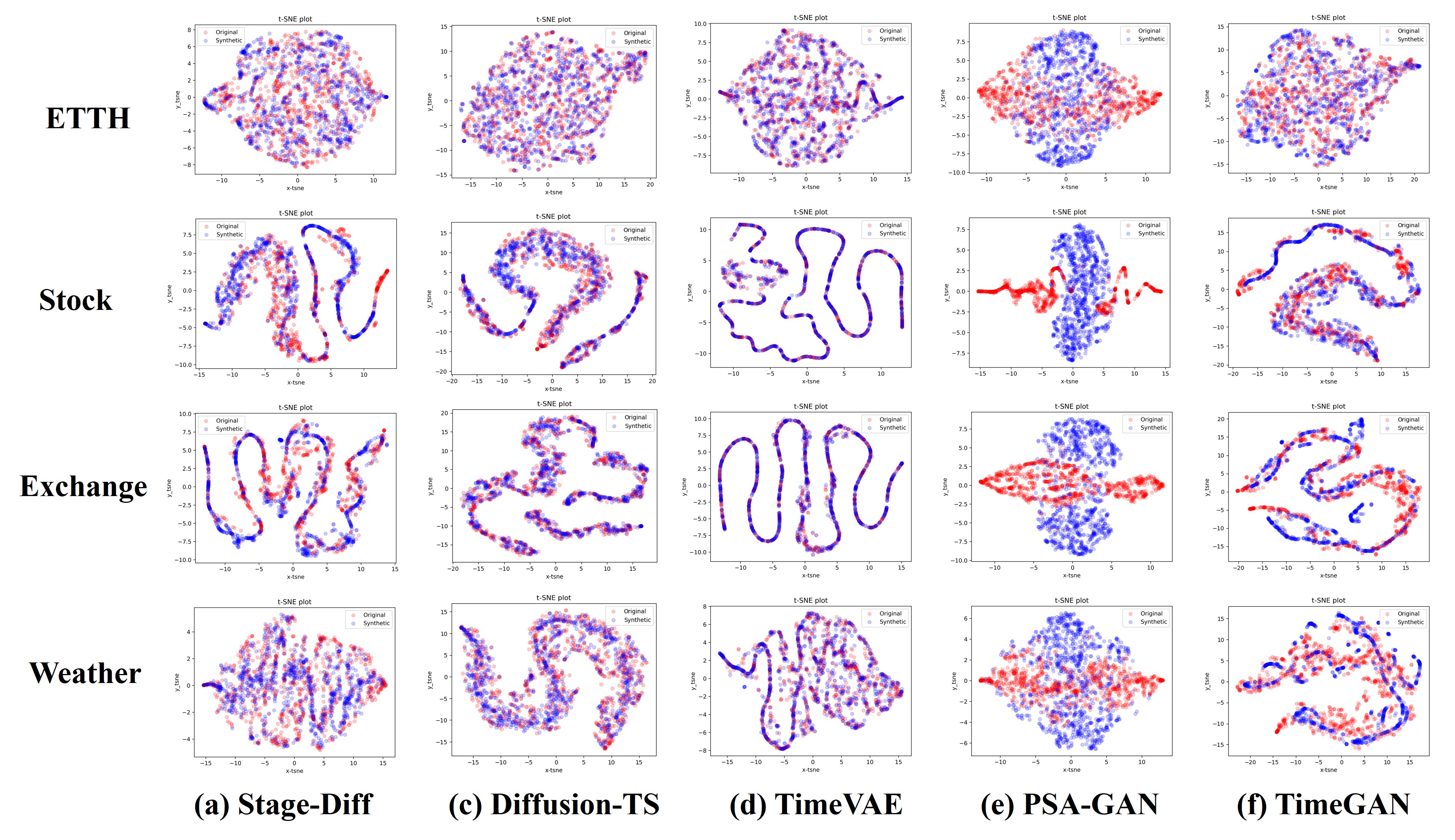}
  \caption{t-SNE visualization. Red denotes original data, and blue denotes synthetic data.}
  \label{fig:result}
\end{figure*}

\section{Experiments}
\subsection{Experiment Setup}
\textbf{Dataset.} We conducted experiments on four publicly available time series datasets: \textbf{1)ETTH} \cite{25}(Electricity Transformer Temperature-hourly), \textbf{2)Stock} \cite{1-timegan} (Google Stock data-daily), \textbf{3)Exchange} \cite{29} Excahnge rate data-daily, \textbf{4)weather} \cite{23} Weather data-minutely. These datasets have been widely used in various time series analysis tasks.

\textbf{Baseline.} We selected the following four time series generation models as baselines: \textbf{1)Diffusion-TS} \cite{18}, \textbf{2)TimeVAE} \cite{13-timevae}, \textbf{3)PSA-GAN} \cite{11}, \textbf{4)TimeGAN} \cite{1-timegan}.

\textbf{Evaluation Metrics.} This paper adopts the synthetic data evaluation metrics from the study \cite{1-timegan},  which mainly include the following three evaluation metrics: 
\begin{enumerate}
\item  \textbf{Visualization}: Using the t-SNE\cite{30}, multivariate time series can be mapped to a two-dimensional space for visualizing the distribution of real and synthetic data. By comparing the overlap between the two distributions, the quality of the synthetic data can be assessed.
\item  \textbf{Discriminative Score}: Used to measure the similarity between the original and synthetic data. A posterior classifier based GRU is employed to differentiate between real and synthetic data. The discrimination score is defined as the absolute difference between the classification accuracy and 0.5.  A lower discrimination score indicates higher quality of the synthetic data.
\item  \textbf{Prediction Score}: The main idea is that synthetic data should exhibit the same performance as the original data when facing the same prediction task. Train a GRU-based predictor on the synthetic dataset and test it on the real dataset. A smaller prediction error indicates higher quality of the synthetic data.
\end{enumerate}

\begin{table*}[h]
\centering
\caption{Results on Multiple Time-Series Dataset (Bold indicates best performance, underscore indicates suboptimal results.)}
\label{tab:result1}
\resizebox{\textwidth}{!}{
\begin{tabular}{c|c|ccccc|ccccc}
\toprule

\multicolumn{2}{c|}{\textbf{Metric}}  & \multicolumn{5}{c|}{\textbf{Discriminative Score}} & \multicolumn{5}{c}{\textbf{Predictive Score}} \\ 
\midrule
\textbf{Dataset} & \textbf{Length} & \textbf{Stage-Diff} & \textbf{Diff-TS} & \textbf{TimeVAE} & \textbf{PSA-GAN} & \textbf{TimeGAN} & \textbf{Stage-Diff} & \textbf{Diff-TS} & \textbf{TimeVAE} & \textbf{PSA-GAN} & \textbf{TimeGAN} \\

\midrule
\multirow{4}{*}{ETTH} & 24  & 0.084 & \underline{0.083} & \textbf{0.014} & 0.497 & 0.106 & 0.132 & \underline{0.122} & \textbf{0.094} & 0.253 & 0.132\\
                      & 64  & 0.093 & \underline{0.092} & \textbf{0.085} & 0.493 & 0.205 & \textbf{0.114} & 0.151& \underline{0.117} & 0.246 & 0.124\\
                      & 128 & \textbf{0.085} & \underline{0.121} & 0.197 & 0.489 & 0.391 & \textbf{0.143} & \underline{0.155} & 0.158 & 0.351 & 0.251\\
                      & 256 & \textbf{0.108} & \underline{0.187} & 0.233 & 0.493 & 0.412 & \textbf{0.152} & \underline{0.223} & 0.249 & 0.332 & 0.309\\
\midrule
\multirow{4}{*}{Stock} & 24  & 0.093 & \underline{0.092} & \textbf{0.015} & 0.469 & 0.102 & \underline{0.037} & 0.046 & \textbf{0.017} & 0.061 & 0.038 \\
                       & 64  & \underline{0.095} & 0.103 & \textbf{0.089} & 0.488 & 0.283 & \underline{0.053} & 0.055 & \textbf{0.051} & 0.057 & 0.139\\
                       & 128 & \textbf{0.101} & \underline{0.138} & 0.140 & 0.482 & 0.443 & \textbf{0.062} & 0.092 & \underline{0.089} & 0.143 & 0.157 \\
                       & 256 & \textbf{0.113} & \underline{0.152} & 0.193 & 0.495 & 0.457 & \textbf{0.069} & \underline{0.103} & 0.152 & 0.209 & 0.314\\
\midrule
\multirow{4}{*}{Exchange} & 24  & \underline{0.104} & 0.121 & \textbf{0.014} & 0.499 & 0.209 & \underline{0.043} & 0.053 & \textbf{0.021} & 0.331 & 0.051\\
                          & 64  & \underline{0.111} & 0.143 & \textbf{0.084} & 0.374 & 0.388 & \textbf{0.082} & 0.094 & \underline{0.089} & 0.391 & 0.173 \\
                          & 128 & \textbf{0.112} & \underline{0.136} & 0.156 & 0.384 & 0.388 & \textbf{0.101} & \underline{0.137} & 0.185 & 0.384 & 0.277\\
                          & 256 & \textbf{0.153} & \underline{0.212} & 0.274 & 0.492 & 0.499 & \textbf{0.092} & \underline{0.177} & {0.321} & 0.471 & {0.308} \\
\midrule
\multirow{4}{*}{Weather}  & 24  & 0.221 & \underline{0.187} & \textbf{0.003} & 0.499 & 0.412 & {0.005} & \underline{0.004} & {0.005} & 0.484 & \textbf{0.002} \\
                          & 64  & \underline{0.193} & 0.195 & \textbf{0.023} & 0.499 & 0.434 & \textbf{0.018} & \underline{0.026} & 0.093 & 0.539 & {0.046} \\
                          & 128 & \textbf{0.164}  & \underline{0.209} & 0.261 & 0.436 & 0.487 & \textbf{0.075} & \underline{0.082} & {0.115} & 0.475 & 0.153 \\
                          & 256 & \textbf{0.189} & \underline{0.276} & 0.387 & 0.443 & 0.499 & \textbf{0.154} & \underline{0.244} & {0.249} & 0.553 & 0.272 \\
\bottomrule
\end{tabular}
}
\end{table*}

\begin{table*}[h!]
\centering
\caption{Results of Ablation Experiments (Bold indicates best performance)}
\label{tab:result2}
\resizebox{0.88\textwidth}{!}{
\begin{tabular}{c|c|cccc|ccccc}
\toprule
\multicolumn{2}{c|}{\textbf{Metric}}  & \multicolumn{4}{c|}{\textbf{Discriminative Score}} & \multicolumn{4}{c}{\textbf{Predictive Score}} \\ 
\midrule
 \textbf{Dataset} & \textbf{Length} & \textbf{w/o CI} & \textbf{w/o CD} & \textbf{w/o Stage} & \textbf{Stage-Diff} & \textbf{w/o CI} & \textbf{w/o CD} & \textbf{w/o Stage} & \textbf{Stage-Diff} \\ 
\midrule
\multirow{4}{*}{ETTH}    & 24   & 0.103  & 0.092   & \textbf{0.071} & 0.084      & 0.136  & 0.134   & \textbf{0.103} & 0.132      \\ 
                        & 64  & 0.152  & 0.101   & \textbf{0.087} & 0.093      & 0.142  & 0.133   & 0.121      & \textbf{0.114}      \\ 
                         & 128    & 0.122  & 0.092   & 0.147      & \textbf{0.085} & 0.193  & 0.187   & 0.198      & \textbf{0.143}      \\ 
                         & 256  & 0.271  & 0.138   & 0.212      & \textbf{0.108} & 0.202  & 0.193   & 0.241      & \textbf{0.152}      \\ 
                         \midrule
 \multirow{4}{*}{Stock}   & 24   & 0.094  & 0.093   &\textbf{0.079} & 0.093      & 0.040  & 0.041   & \textbf{0.013} & 0.037      \\ 
                       & 64    & 0.137  & 0.124   & 0.109      & \textbf{0.095} & 0.088  & 0.063   & 0.152      & \textbf{0.053}      \\ 
                        & 128  & 0.133  & 0.113   & 0.122      & \textbf{0.101} & 0.102  & 0.083   & 0.143      & \textbf{0.062}      \\ 
                       & 256 & 0.138  & 0.129   & 0.215      &\textbf{0.113} & 0.142  & 0.135   & 0.182      & \textbf{0.069}      \\
                       \midrule
\multirow{4}{*}{Exchange}& 24     & 0.118  & 0.111   & 0.105      & \textbf{0.104} & 0.051  & 0.047   & \textbf{0.028}      & 0.043      \\ 
                        & 64   & 0.133  & 0.124   & 0.155      & \textbf{0.111} & 0.103  & 0.095   & 0.092      & \textbf{0.082}      \\
                        & 128    & 0.189  & 0.163   & 0.232      &\textbf{0.112} & 0.206  & 0.132   & 0.213      & \textbf{0.101}      \\
                       & 256 & 0.233  & 0.201   & 0.318      & \textbf{0.153} & 0.194  & 0.173   & 0.237      & \textbf{0.092}      \\ 
                       \midrule
\multirow{4}{*}{Weather} & 24 & 0.314  & 0.307   & 0.268      & \textbf{0.221} & 0.005  & 0.004   & \textbf{0.002} & 0.005      \\ 
                        & 64      & 0.214  & \textbf{0.173}   & 0.209      & {0.193} & 0.020  & \textbf{0.018}   & 0.026      & \textbf{0.018}      \\ 
                        & 128  & 0.320  & 0.203   & 0.392      & \textbf{0.164} & 0.176  & 0.133   & 0.204      & \textbf{0.075}      \\ 
                        & 256       & 0.283  & 0.234   & 0.302      &\textbf{0.189} & 0.259  & 0.209   & 0.397      & \textbf{0.154}      \\
\bottomrule
\end{tabular}
}
\end{table*}

\subsection{Experiment Results}
\textbf{Visualization.} The experimental results are depicted in Figure \ref{fig:result}. Stage-Diff, Diffusion-TS and TimeVAE generate synthetic time-series data that completely aligns with the distribution of real data. PSA-GAN achieves partial alignment between synthetic and real data distributions in the central regions but fails to achieve complete coverage. This could be due to the loss of temporal patterns during the transition from coarse-grained to fine-grained modeling. TimeGAN demonstrates strong performance on the ETTH, Stock, and Exchange datasets, but suffers from performance degradation on the Weather dataset, likely due to its high dimensionality.

\textbf{Discriminative Score and Prediction Score.} 
The experimental results, as shown in the table\ref{tab:result1}, demonstrate that Stage-Diff achieves optimal or suboptimal performance across different datasets with varying sequence lengths. Moreover, thanks to the stage-wise generation approach, its performance remains stable as the sequence length increases. Diff-TS achieved a performance second only to Stage-Diff. TimeVAE performs well on shorter sequences but suffers significant performance degradation as the sequence length grows. PSA-GAN and TimeGAN, both GAN-based models, exhibit unstable performance due to the inherent instability of adversarial training, with the instability becoming more pronounced as the sequence length increases. 


\textbf{Ablation Experiments.} 
The performance gain of Stage-Diff mainly comes from its stage-wise time series generation, channel-independent modeling within each stage and multi-channel fusion modeling between stages.  To verify their contributions, we designed three variants of Stage-Diff: \textbf{w/o CI} removes channel-independent modeling within stages, \textbf{w/o CD} removes multi-channel fusion during inter-stage information transmission, and \textbf{w/o stage} replaces the stage-wise generation with global modeling. To compensate for the loss of multi-channel fusion modeling caused by the removal of inter-stage information transmission, an additional layer of multi-channel convolution is integrated into the decomposition process.

The experimental results, as shown in the table \ref{tab:result2}, indicate that \textbf{w/o stage} performs well for shorter sequence lengths, suggesting that the stage-wise generation approach is more suitable for long time series, while for shorter sequences, it may reduce the model's ability to capture sequence dependencies. The performance of \textbf{w/o CI} is overall inferior to \textbf{w/o CD}, indicating that the performance gain from channel-independent modeling is greater than that of channel fusion modeling in the time series generation process. However, the effective integration of both contributes to achieving even better results. Overall, the complete Stage-Diff model consistently achieves the best results, demonstrating the effectiveness of the three design components for time series generation tasks.

\section{Conclusion}
In this paper, we propose a stage-based long-term time series generation method, Stage-Diff, based on a diffusion model. By repeatedly combining intra-stage sequence generation and inter-stage information transmission, the model is extended to longer time scales. Within each stage, channel-independent modeling is employed to capture dependencies at different time scales through progressive decomposition. Across stages, multi-channel historical information at various scales is fused and passed to the next time stage, enabling effective capture of both intra-channel and inter-channel relationships. The transmission of multi-channel historical information ensures the long-term dependency of the sequences, while its integration with stage-based generation provides a potential solution for modeling data distribution shifts. Extensive experiments on multiple real-world datasets validate the effectiveness of the proposed model, and ablation studies further demonstrate the contribution of key modules. In the future, we will explore the completeness of the distribution of synthetic time series to generate more practical synthetic time series.



\end{sloppypar}
\end{document}